\documentclass[11pt, a4paper, logo, twocolumn, copyright]{googledeepmind}

\usepackage[authoryear, sort&compress, round]{natbib}
\usepackage{subfiles}
\usepackage{subcaption}

\title{Latent Space Communication via K-V Cache Alignment}

\correspondingauthor{ldery@google.com}


\renewcommand{\today}

\author[1]{Lucio M. Dery}
\author[1]{Zohar Yahav}
\author[1]{Henry Prior}
\author[1]{Qixuan Feng}
\author[1]{Jiajun Shen}
\author[1]{Arthur Szlam}

\affil[1]{Google DeepMind}

\begin{abstract}
Solving increasingly complex problems with large language models (LLMs) necessitates a move beyond individual models and towards multi-model systems that can effectively collaborate. While text has traditionally served as the medium for inter-model communication, a richer and more efficient exchange is possible if models can access each other's internal states directly. In this paper, we propose learning a shared representation space that aligns the k-v caches of multiple models, creating a high-bandwidth channel for collaboration without altering the underlying pre-trained parameters. We do so by  augmenting each model with adapters to translate its state into and out of this shared space. Via a suite of experiments with Gemma-2 models, we demonstrate that this approach not only enables seamless inter-model communication but also improves individual model performance. We also show that the shared space allows for the direct transfer of learned skills, such as soft prompts, between different models. Our work represents a significant step towards a future where models can fluidly share knowledge and capabilities.
\end{abstract}

\begin{document}

\maketitle

\section{Introduction}
\looseness=-1 As the abilities of frontier large language models (LLMs) continue to expand, our community is rapidly producing many models with a diverse range of both generalist and domain-specific capabilities \citep{brown2020language, chowdhery2023palm, comanici2025gemini}. This growing pool of models presents both an opportunity and a challenge: how can we effectively orchestrate these individual models to solve problems that are beyond the scope of any single one? Such collaboration between models can take two primary forms. The first is multi-model systems, such as LLM agents \citep{wang2024openhands, lu2024ai}, model cascades \citep{yue2023large, kolawole2024revisiting}, model frankensteins \citep{akiba2025evolutionary} and mixture-of-expert models \citep{filippova2024no}. The second is skill reuse \citep{sabry2024assessing}, where modules or knowledge from one model -- like soft prompts \citep{lester2021power} or prefix k-v caches \citep{li2021prefix} -- are shared and leveraged by others.

\looseness=-1 Towards the north-star of seamless inter-model collaboration \citep{kandpal2023git}, we provide evidence showing the feasibility of allowing models to read and write to each other's key-value (k-v) cache latent space. 
A transformer's \citep{vaswani2017attention} k-v cache is a rich, natural representation of its internal state. A sufficiently high dimensional cache vector space could allow for much higher communication bandwidth between models than token-level text exchange
\citep{hao2024training, moschella2022relative}. By accessing each other's latent space, models could directly read each others' state and extract relevant knowledge, follow (soft) reasoning chains, and reuse partial computations or skills, leading to a more efficient collaboration on challenging tasks. 

\begin{figure}[t!]
\includegraphics[scale=0.25]{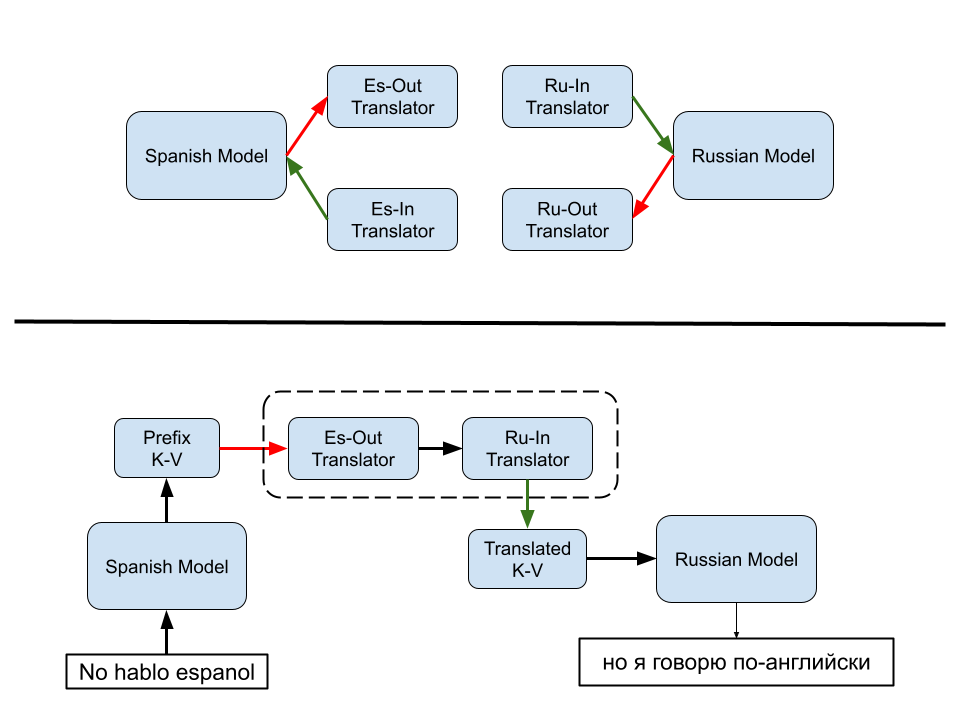}
\caption{Overview of an example instantiation of our framework. [Upper] Each model (here Russian and Spanish LMs) is augmented with adapters for translating a k-v cache block into and out of the shared latent space. E.g Es-Out translates a k-v cache out of the Es model's latent space and into the shared space. [Lower] We can perform mixed language modelling (for context switching settings), by translating generated prefix caches from a source model and continuing to decode from the target model.}
\label{fig:mt-translation-arch}
\centering
\end{figure}
Unfortunately, even given the same 
input text, we cannot expect the latent spaces of two models trained under different conditions to be interchangeable \citep{ainsworth2022git}. This is because any given token passing through a model creates conditional dependencies on the k-v space of the layer(s) before it, causing the discrepancies in latent space between any pair of models with different parameters to compound with depth \citep{nguyen2020wide, friedman2023comparing}.

Taking inspiration from the machine translation literature \citep{richens1956general, lu2018neural}, given a set of models with possibly disjoint latent spaces, we propose to learn a global, shared k-v cache representation space that any model in the pool can read from and write to. Specifically, we augment each model with two adapters: one for translating its k-v cache block into the shared representation space, and another for translating a given shared space cache block into its model-specific latent space (Figure \ref{fig:mt-translation-arch}). As designed, this is an implicit (shared) latent space and presents a scalable communication channel with parameter overhead that only grows linearly with the number of models in the pool.

Through a battery of experiments on Gemma-2 \citep{team2024gemma} open models, we demonstrate 
that we can teach a set of language models trained under a range of different conditions to read and write to each other's k-v cache latent space. To achieve this, we provide a scalable architecture and efficient algorithm for learning a global latent space without modifying the individual models' pre-trained parameters.
We share the following findings:
\begin{enumerate}
    \item Given a set of language models that differ along axes such as size, training data distribution and random initialization, we are able to learn a global shared k-v cache latent representation space.
    \item Given sufficient data for learning the global latent space, we can improve the individual models' language modelling performance by passing prefix k-v caches through the shared space before modelling the suffix text.
    \item A global shared latent space enables learned modules to be portable between models. Specifically, we conduct experiments where soft-prompts, learned on a particular task for one model, can be translated via our global latent space, and used directly by another model to perform the same task. This allows modules learned by individual models to become a shared resource of the whole pool.
\end{enumerate}
\section{Methodology}
We enhance a pool of models with the ability to read and write to each other's k-v cache representation spaces. By \textit{read}, we mean a model can translate from another's k-v cache space into its own, and \textit{write} implies a model translating its k-v cache into another model's representation space.  We do so via a mechanism where individual model latent spaces are mediated by a learned globally shared, but implicit, latent space. In this section, we will detail our framework with architecture specifics and a learning algorithm. 

\subsection{Setup}
\looseness=-1 Consider a pool of N models indexed as $\{m_1, … , m_i, …, m_N\}$. We would like to learn a shared latent space, $\mathbf{\Sigma}$, amongst these models parameterized as follows:
for each model, we learn a map  $\mathbf{T}^{[m_i \rightarrow \mathbf{\Sigma}]}$ into the shared space and another map $\mathbf{T}^{[\mathbf{\Sigma} \rightarrow m_i]}$ out of the space. Thus, assuming that we have the cache for model $m_i$, $KV^{i} \in \mathbf{R}^{B\times S \times L_i \times D_i}$, our parameterization instantiates $\mathbf{T}^{[m_i \rightarrow \mathbf{\Sigma}]}:  \mathbf{R}^{B\times S \times L_i \times D_i} \rightarrow  \mathbf{R}^{B\times S \times Q} $ and  $\mathbf{T}^{[\mathbf{\Sigma} \rightarrow m_i]}:  \mathbf{R}^{B\times S \times Q} \rightarrow \mathbf{R}^{B\times S \times L_i \times D_i }$. Where B, L, S, D are batch, layer, sequence and model dimensions respectively. For simplicity, we assume that all the models in the pool share the same vocabulary and so for a piece of text, $S$ is the same for a for all models. Our method does not require this assumption, but it is convenient for conveying a simple setup.

Though each model may have a different number of layers $L_i$ and model dimension $D_i$, we map all of them to a shared embedding space of fixed size, $Q$. The shared space therefore has no direct layer-wise demarcation. Instead of forcing a pre-specified assignment per-model, we leave it to the output maps to learn to appropriately reconstruct the layer demarcations. Each model's map into the global shared space mixes layer-wise information ($L_i \times D_i \rightarrow Q$) whilst the mapping out of the shared space performs a model-specific reconstruction of layer-wise information ($Q \rightarrow L_i \times D_i $). We prefer such a design because layers might not have one-to-one correspondence across models of different sizes and even if a pair of models have the same number of layers, it is not guaranteed that they map directly to each other due to the presence of residual connections \citep{nguyen2020wide}. 

We highlight that each model has an adapter \footnote{We will use the terms adapter, mapping and translator interchangeably} into and an adapter out of the shared space. Since the shared space is global, once a model translates into the shared space, that latent representation can be translated into the local latent space of any of the other models using the corresponding adapter (including the original model itself). This ensures that the number of parameters required to learn a shared space only grows linearly as the number of models. Another advantage of this design is that it is easy to incorporate new models into the pool. Adding a new model, $m_k$, would simply involve learning the pair $\mathbf{T}^{[m_k \rightarrow \mathbf{\Sigma}]}$ and $\mathbf{T}^{[\mathbf{\Sigma} \rightarrow m_k]}$ whilst keeping the maps of all the other models (and the shared space itself, $\mathbf{\Sigma}$) fixed.

\begin{figure}[t!]
\includegraphics[scale=0.245]{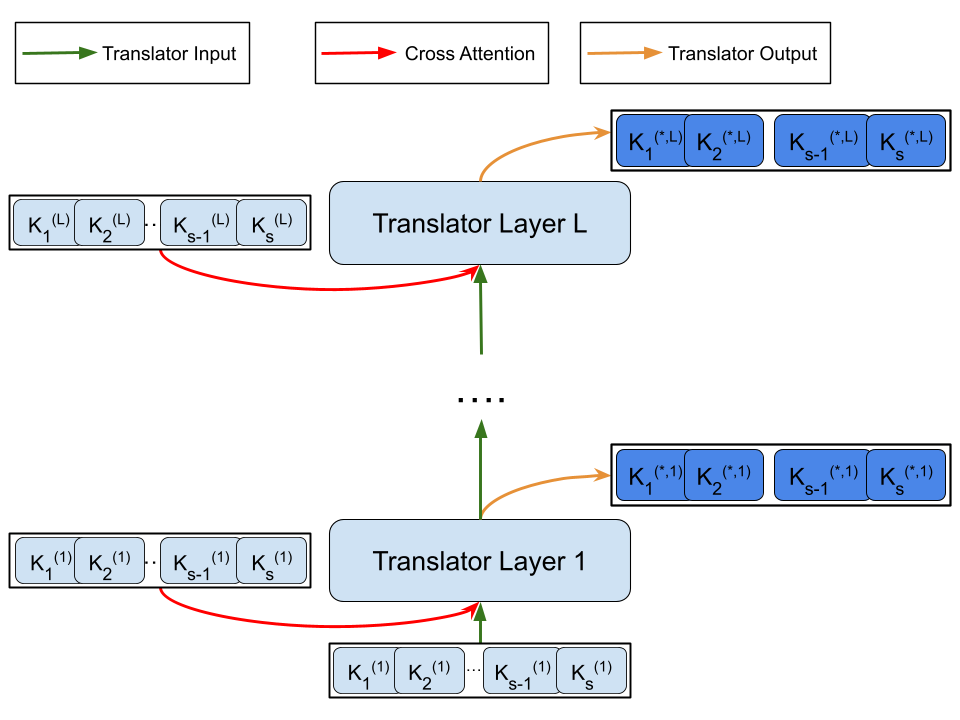}
\caption{Cross attention architecture for mapping into and out of the shared latent space. The first \textit{layer} k-or-v cache serves as seed input to the model. Each layer in the module cross-attends to the corresponding layer in the input k-or-v (not both at the same time) block, meaning that the number of layers in the translator is equal to the number of layers in the corresponding model. We use the superscript $^{*}$ to represent vectors in $\mathbf{\Sigma}$.}
\label{fig:cross_attn}
\centering
\end{figure}
\subsection{Translator Architecture}
\label{subsec:trans_arch}
Given that the relationship between k-v caches of different models may be highly non-linear, we parameterize the transformations into and out of the shared embedding space with multi-layer transformer models. For any model $m_i$ in our pool, we use the same high level architecture for both $\mathbf{T}^{[m_i \rightarrow \mathbf{\Sigma}]}$ and $\mathbf{T}^{[\mathbf{\Sigma} \rightarrow m_i]}$:
\begin{enumerate}
    \item \textbf{Input Transformation:} Consider an input cache block (either key or value) $KV$. For $\mathbf{T}^{[m_i \rightarrow \mathbf{\Sigma}]}$, this block will have dimensions $KV^{\text{local}} \in \mathbf{R}^{B\times S \times L_i \times D_i}$ but for $\mathbf{T}^{[\mathbf{\Sigma} \rightarrow m_i]}$, which receives a block from the global shared space, $KV^{\text{global}} \in \mathbf{R}^{B\times S \times Q}$. To keep the architecture symmetric we perform a pre-processing step of reshaping $\mathbf{R}^{B\times S \times Q} \rightarrow \mathbf{R}^{B\times S \times L_i \times (Q // L_i)}$ when translating from global to local space with $\mathbf{T}^{[\mathbf{\Sigma} \rightarrow m_i]}$. This requires adding a constraint that $\{\forall L \in L_i, \; Q \mod L = 0\}$ so the global space is the same dimension for all models.

    We now apply a simple non-linear transformation consisting of a layer-normalization \citep{ba2016layer}, linear transform and a GELU activation \citep{hendrycks2016gaussian}. The resulting intermediate result has dimensions $\mathbf{R}^{ B \times S \times L_i \times D^{\prime}_i}$ where $D^{\prime}_i$ is the output dimension of the linear transform. \footnote{Our choice of linear transform preserves $L_i$ but we are free to expand or contract $L_i \rightarrow \tilde{L}_i$. Using $\tilde{L}_i < L_i$ can reduce the latency of running the adapter but at the cost of expressiveness.} Note that we differentiates between key and value caches by using different sets of parameters for the linear transform.
    \item \textbf{Cross Attention:} The main workhorse of the latent space adapter is the cross attention module. Figure \ref{fig:cross_attn} is a pictorial representation of the architecture. Using this layer-wise cross-attention architecture allows us to model the hierarchical process that generated the k-v cache block in the first place.
    For this module, we share parameters between key and value type cache input blocks. At a given layer, we use the output of the previous layer to generate the query vector whilst the key and value vectors are generated by the corresponding k/v cache designated for cross-attention.
    \item \textbf{Output Transformation:} We concatenate the outputs of the cross-attention layers together (along the last dimension) to obtain $\mathbf{R}^{B\times S \times ( L_i \times D^{\prime}_i)}$ and pass this through a non-linear output transformation (exact same structure as input transformation described above) to output a cache block in $\mathbf{R}^{B\times S \times ( L_i \times D_i)}$. This is then reshaped to produce $KV^{\text{local}} \in \mathbf{R}^{B\times S \times L_i \times D_i}$ as the final output.
\end{enumerate}

\subsection{Learning Algorithm}
\label{subsection:learning_algo}
We explore two losses to provide signal for learning the global shared latent space.

\subsubsection{Reconstruction Loss}
We consider a reconstruction loss for converting the k-v cache of one model to another. Let $KV_{x}$ denote a model's k-v cache corresponding to a piece of text, x.
$$\mathcal{L}_{\mathrm{recon}} = \sum_{m_i, m_j} \big\| \mathbf{T}^{[\mathbf{\Sigma} \rightarrow m_j]}\big(\mathbf{T}^{[m_i \rightarrow \mathbf{\Sigma}]}(KV^{i}_{x})\big) -  KV^{j}_{x} \big\|^{2} $$
The equation above shows a translation between models $m_i$ and $m_j$ which passes through the $\mathbf{\Sigma}$ only once. In theory, we can also have chains of translations in and out of the shared space  ($\mathbf{T}^{[\mathbf{\Sigma} \rightarrow m_j]} \ldots \mathbf{T}^{[m_{l}] \rightarrow \mathbf{\Sigma}} \odot \mathbf{T}^{[\mathbf{\Sigma} \rightarrow m_{l}]} \odot \mathbf{T}^{[m_{l - 1}] \rightarrow \mathbf{\Sigma}} \odot \mathbf{T}^{[\mathbf{\Sigma} \rightarrow m_{l-1}]} \ldots  \mathbf{T}^{[m_i \rightarrow \mathbf{\Sigma}]}$) linking the two models at the ends of the chain. However, back propagation through this loss would be computationally expensive.

\subsubsection{Suffix Language Modelling}
Even small reconstruction errors can cascade into large prediction errors. Also, an exact reconstruction of the target model's k-v cache means that we are upper bounded by its original language modelling performance.  Since we are primarily concerned with language models, we propose a suffix language modelling loss that conditions on a prefix k-v cache translated from another model. Specifically, given a text sequence, x, of length $\tau$, and a source-target model pair $(m_i, m_j)$, we use our cache-translation on an  $s$-length chunk of generated k-v cache: $KV^{i \rightarrow j}_{x[:s]} = \mathbf{T}^{[\mathbf{\Sigma} \rightarrow m_j]}\big(\mathbf{T}^{[m_i \rightarrow \mathbf{\Sigma}]}(KV^{i}_{x[:s]})\big)$. We then use this chunk as a prefix for language modelling the remaining $\tau$-s tokens by the target model: 
$$\mathcal{L}_{\mathrm{LM}} = \sum_{m_i, m_j} CE\bigg(x_{[s:\tau]}, m_j\big(KV^{i \rightarrow j}_{x[:S]}\big)_{[s \rightarrow \tau]} \bigg) $$

CE is the cross entropy loss and $m_j\big(\cdot \big)_{[s \rightarrow \tau]}$ is the $\tau$-s length suffix output of model $m_j$ given an input k-v cache prefix. Note that since each model involved in the suffix language modelling loss operates on disjoint sections of the text, we can apply it to models that have different vocabularies. However, the reconstruction loss can not be directly applied to models with disjoint vocabularies unless we a-priori define a mapping between the two vocabularies.



\section{Experimental Setting}
\textbf{Pre-trained Models:} For our experiments, we pre-train a range of Gemma-2 \citep{team2024gemma} models of sizes between 100M-400M parameters (not counting the token embeddings). For all our experiments, we keep the pre-trained models frozen and only learn the adapters into and out of the latent space. We use the multilingual C4 dataset \citep{2019t5} with a maximum sequence length of 512 tokens per example sequence. Unless otherwise specified, we focus on the English, Russian and Spanish splits of this dataset for all our experiments. \\
\textbf{Adapter architecture Details: }
For the adapters into the shared space, we use the Gemma-2 architecture but modify it to allow for cross-attention. We fix the number of heads to 32 and head dimension to 64 for all experiments. Two hyper-parameters control the size of the adapter models (since it has the same number of layers as the base model): \texttt{embed\_dim\_factor} ($= \frac{D^{\prime}_i}{D_i}$) and \texttt{translation\_dim\_factor} (by what factor the fully connected layers expand the residual stream dimensionality). For all experiments except when we are ablating the adapter model size, we set \texttt{embed\_dim\_factor} = 2 and \texttt{translation\_dim\_factor} = 1. This generally results in each translator being $\sim \frac{1}{4}$ the base model size.\\
\textbf{Training the translator: } We train for 50k steps using learning rate warmup for 2.5k (5\%) steps and cosine decay learning rate schedule \citep{loshchilov2016sgdr}. The initial learning rate value is set to $1e^-6$ before warmup and it is eventually decayed to 0.0 at the end of the training. We use the AdamW optimizer \citep{loshchilov2017decoupled} with the default hyper-parameters in Optax \citep{hessel2020optax} except for the learning rate which we sweep over in the set $[1e^{-3}, 1e^{-4}, 1e^{-5}]$. All gradient norms were clipped to 1.0 before being applied. During training, we sweep batch sizes in the range $[256, 512, 1024]$. Unless otherwise specified, our experiments use the first 128 tokens as prefix and the remaining as suffix for generation with the target model.
\begin{figure*}[ht!]
    \centering
    \begin{subfigure}[t]{0.48\textwidth}
        \includegraphics[scale=0.3]{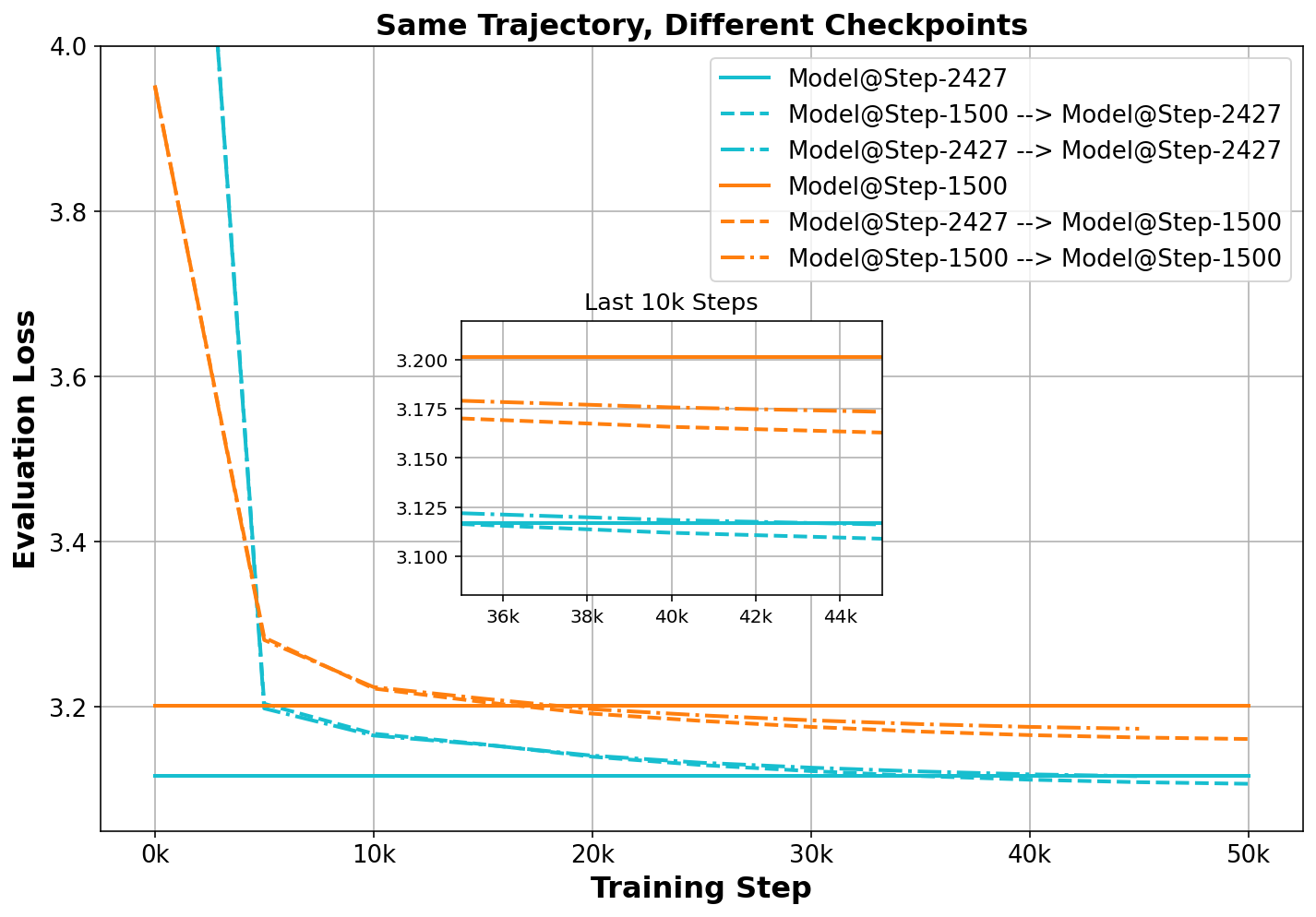}
        \caption{We can improve the performance (on language modelling text suffixes)  of two models that are different checkpoints of the same training trajectory by translating prefixes through the global latent-space. All curves are comparable to each other since the eval sets are the same.}
        \label{fig:diff_chkpt}
    \end{subfigure}
    \hfill
    \begin{subfigure}[t]{0.48\textwidth}
        \includegraphics[scale=0.3]{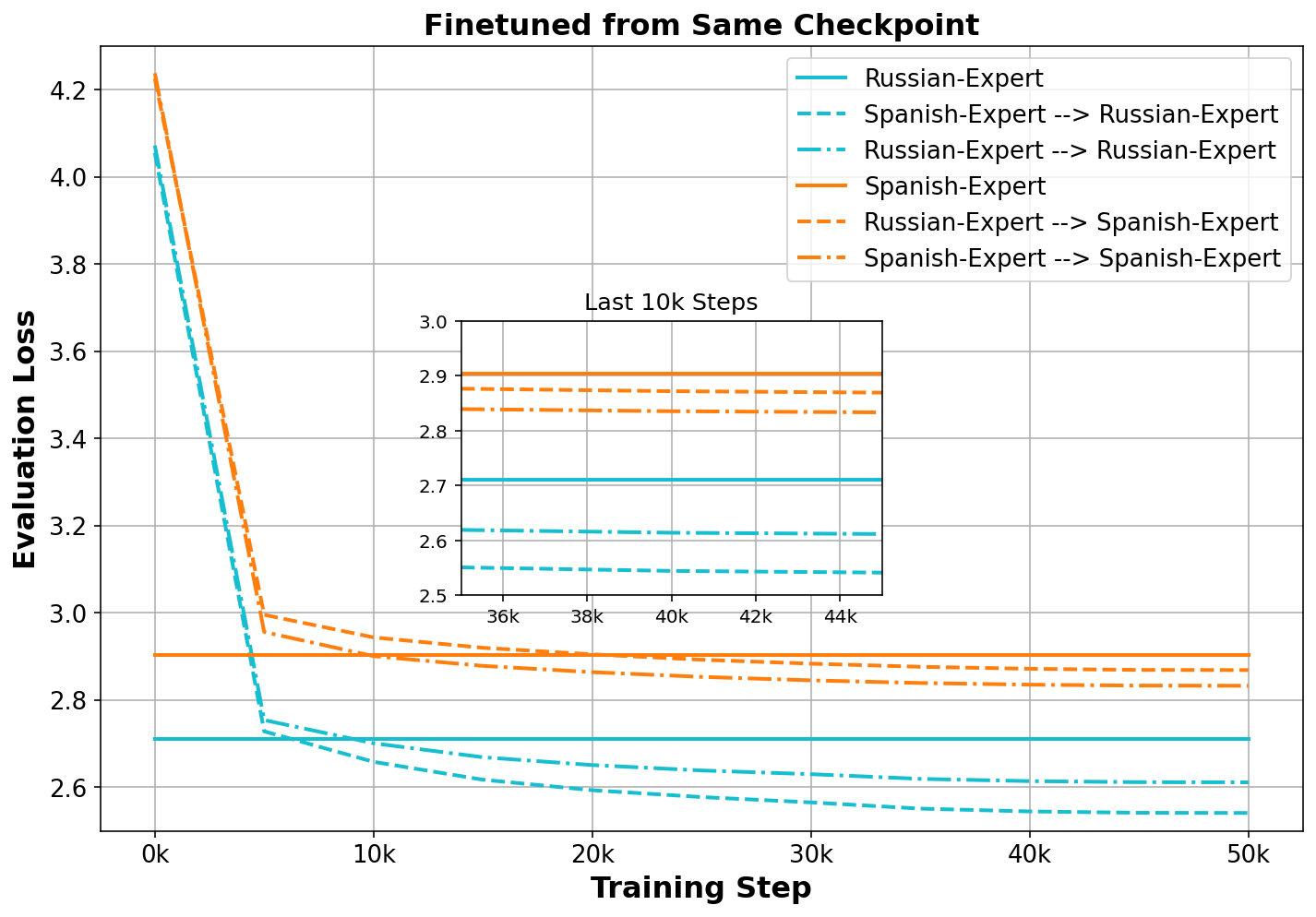}
        \caption{We take the same checkpoint and fine-tune it on different data distributions to produce expert models in Russian and Spanish. We are able to learn a global latent-space between these two models. Only curves of the same color are comparable since each language has a different eval set.}
        \label{fig:finetuned_same_chkpt}
    \end{subfigure}
    
    \medskip
    
    \begin{subfigure}[t]{0.48\textwidth}
        \includegraphics[scale=0.3]{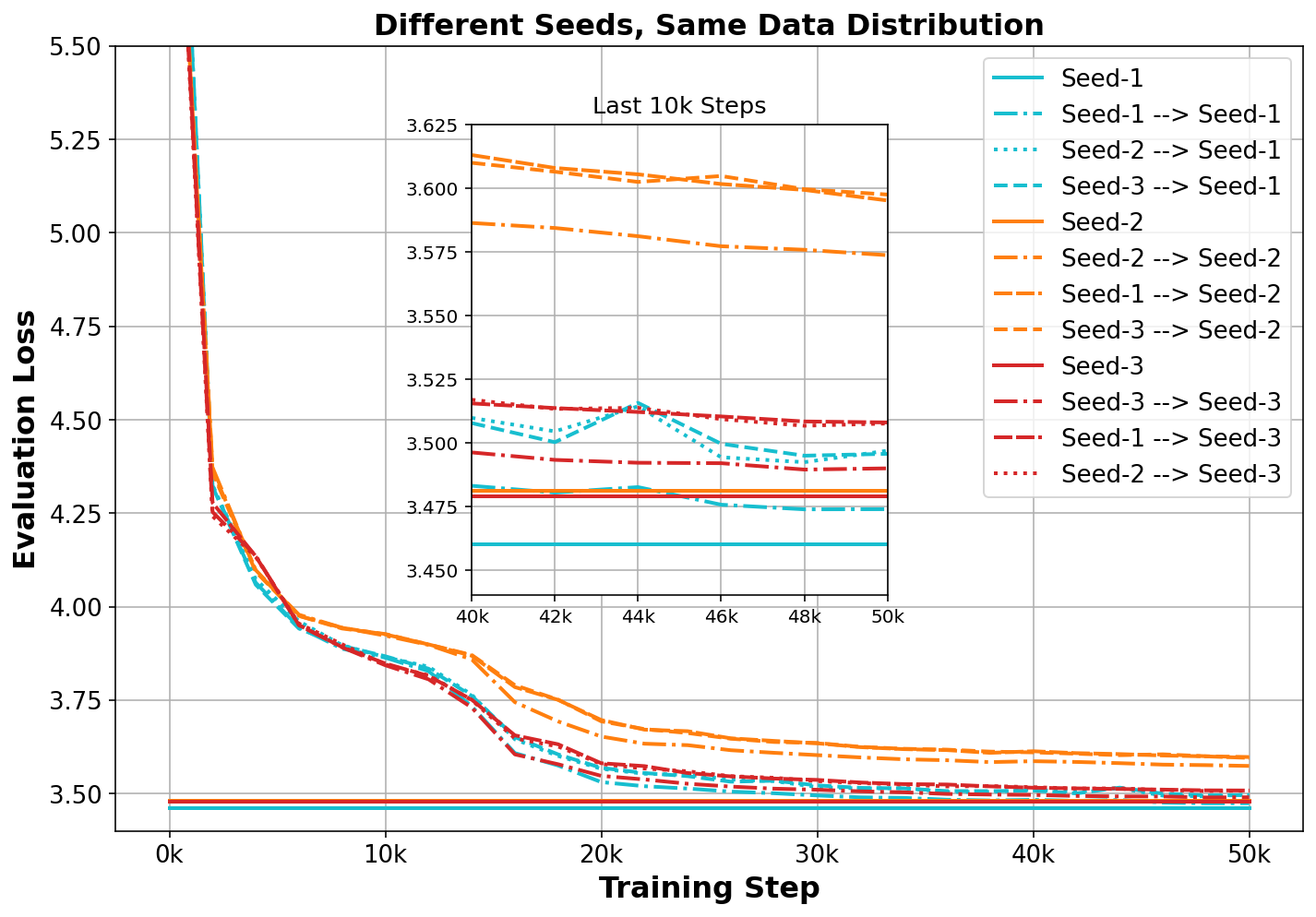}
        \caption{Our method works with pools of models that have different random initialization. We see that all our paths come close to or surpass the base models performances.}
        \label{fig:diff_seeds_3}
    \end{subfigure}
    \hfill
    \begin{subfigure}[t]{0.48\textwidth}
        \includegraphics[scale=0.3]{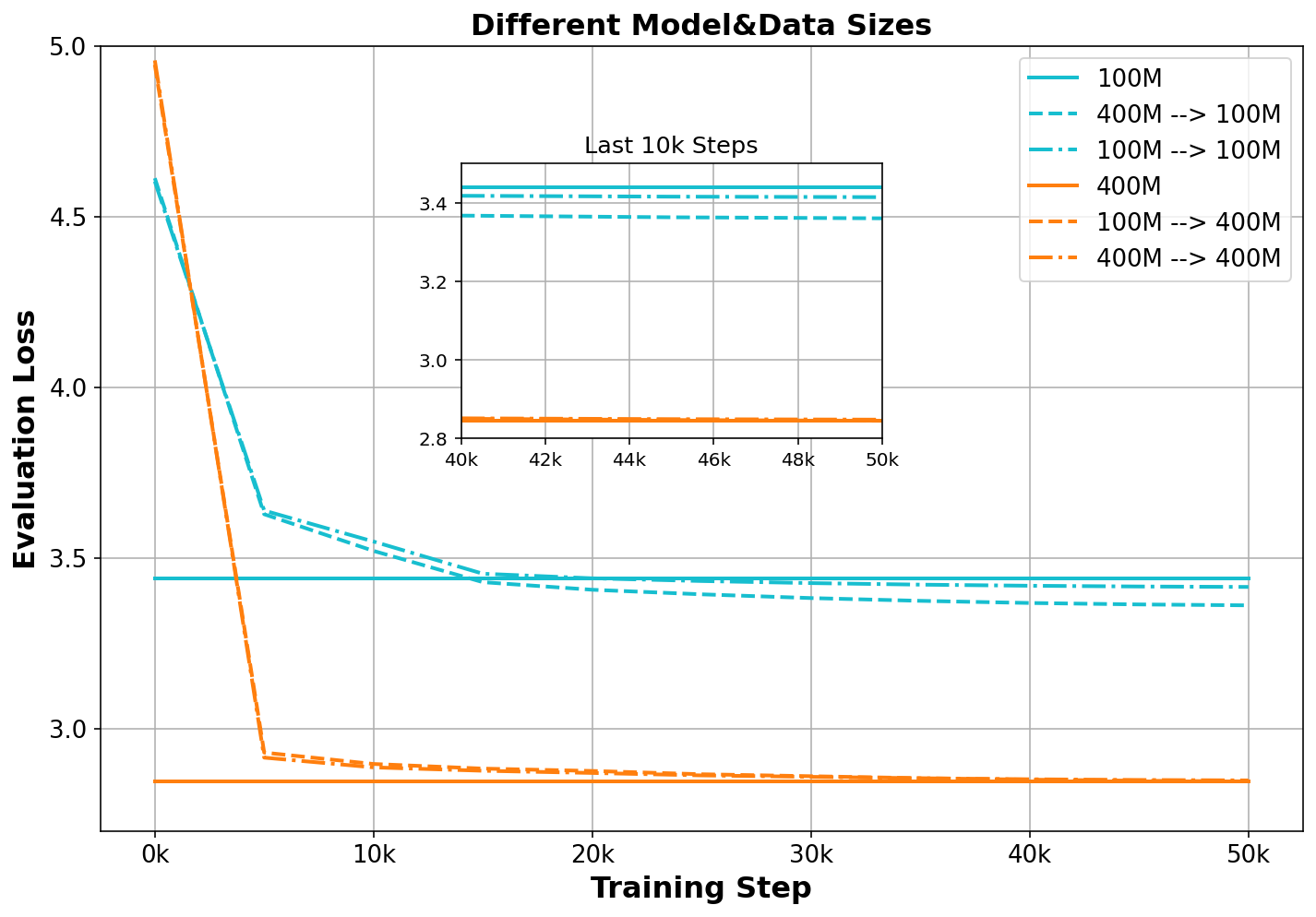}
        \caption{The 400M model has 4$\times$ more layers and is trained on more data than the 100M model. Our framework allows this pair of models to collaborate on the language modelling task despite these differences.}
        \label{fig:diff_model_and_data}
    \end{subfigure}
    \caption{We can learn a shared latent space between different models with different degrees of overlap in their training trajectories. Evaluation loss is negative log likelihood on held-out text.}
    \label{fig:variations}
\end{figure*}

\section{Models can be taught to communicate via a global k-v latent space}
Using our framework, we design a series of experiments to show that it is possible to teach a set of models (that share the same token vocabulary) to cooperate by leveraging a global shared k-v latent space. We consider two main kinds of model sets: models that are fine-tuned or adapted from the same origin and a set of models that have entirely disjoint training trajectories. For all our experiments in this section, unless otherwise specified, we use only the suffix language modelling loss.

\subsection{Same trajectory, different checkpoints}
\label{section:same_traj}
We train a 200M model on a mixed language distribution of 50\% English, and 25\% each of Spanish and Russian till Chinchilla optimal \citep{hoffmann2022training}. We take 2 checkpoints – the last checkpoint and a checkpoint in the middle of training. We consider the former model as a “stronger” checkpoint and primarily seek to improve the language modelling performance on the earlier checkpoint. Since the training trajectory of the earlier checkpoint is entirely contained in the last, we posit that their k-v latent spaces should be highly related, and translating between these models should be reasonably easy.

Figure \ref{fig:diff_chkpt} shows the results of this experiment. We see an improvement beyond the base model performance when the prefix k-v cache is the output of the global shared space. We posit that the latent space sharpens the relevant k-v cache features in a way that improves the end model's task performance. We observe this improvement even in the case where we do a cyclic translation of the model's prefix-kv cache into the global space and back into its local latent space. Figure \ref{fig:diff_chkpt} also suggests that it takes less data to learn a translation from the stronger model's k-v cache space to the weaker model's, as indicated by the training step at which using the global k-v space matches or improves upon the base model performance.

\subsection{Same origin model, different fine-tuning distributions} 
\label{section:diff_ft}
\looseness=-1 We produce two expert models, one in Russian (trained on 100\% Russian data)  and another in Spanish  (trained on 100\% Spanish data) that are branched from the same parent model: the last checkpoint model from Section \ref{section:same_traj} which was trained on the specified mixed language distribution.

We created a Russian-Spanish parallel dataset for this experiment using the English C4 dataset as the pivot source. Using publicly available Gemini API call, we translated a subset of the English text into Russian and Spanish. By pairing the corresponding translations from each language, we formed a direct Russian-Spanish parallel corpus with 10M pairs for the training set and an 800K pairs for the evaluation set.

To illustrate the setup, suppose we are using Russian as the target language and Spanish as the source language. We pass a prefix of Spanish text to the Spanish language expert to obtain a prefix cache, and then generate the Russian suffix text with the Russian language expert after translating the (prefix) cache through the global latent space into the Russsian expert's latent space. We do very weak alignment of the parallel text -- we simply skip the first prefix-length-tokens in the target language and continue decoding.

Though this setting represents a more extreme latent space divergence than we have in Section \ref{section:same_traj}, our approach is similarly capable of learning a good shared latent space as Figure \ref{fig:finetuned_same_chkpt} indicates. Again, when trained on enough data, cyclicly translating a prefix k-v cache through the latent space can actually outperform decoding using the model's untouched k-v cache.

\subsection{Same data distribution but different random initialization}
\label{section:diff_seeds}
For the above sections, the training trajectories of the model pool have had non-zero overlap. Next, we investigate pools that share no trajectory overlap, but are trained on the same data distribution. We use 3 100M models trained on the same mixed language distribution as in Section \ref{section:same_traj} but initialize each model with different random seeds. 

We use a prefix length of 64 tokens and sequence length of 128 for these experiments since 3 models presents a higher memory overhead. As in previous sections, even though these models do not share training trajectories, we are still able to teach them to leverage each other's k-v caches (Figure \ref{fig:diff_seeds_3}).

\subsection{Different Model Sizes}
\label{section:diff_szs}
Figure \ref{fig:diff_model_and_data} shows that we are able to translate the k-v caches of models of different sizes. Specifically, though the 400M model has 16 layers whilst the 100M model has 4 layers, our adapters are able to learn an appropriate mapping. Similar to the earlier experiments in this section, we are able to improve the performance of the weaker 100M model by using prefix k-v caches translated from the stronger 400M model. Note that the two models have disjoint training trajectories not only because of different random initialization, but also due to different amounts of training data (since they are both trained Chinchilla-optimally and thus the 400M is trained with more data than the 100M model).

\section{The global k-v latent space is extensible}
\begin{figure}[h!]
\includegraphics[scale=0.32]{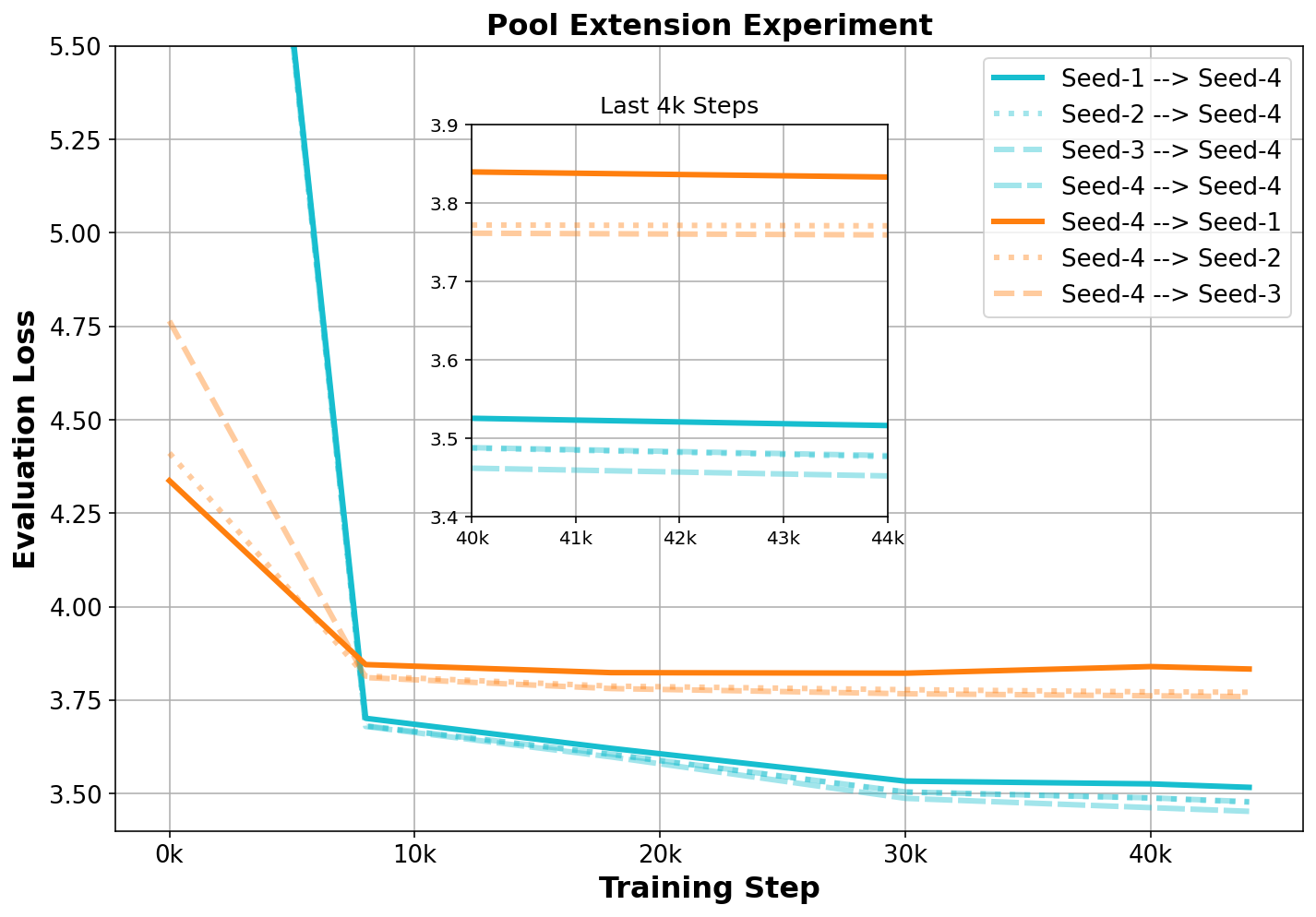}
\caption{We extend a pool of 3 models with different seeds-\{1, 2, 3\} with a fourth model. \textbf{Solid lines mark translation paths that were not trained on}. Without explicitly training on a these translation paths, we are still able to zero-shot translate the between the k-v caches of the two models -- seeds-\{1, 4\} -- with only mild performance degradation compared to fully trained paths.}
\label{fig:pool_extension}
\centering
\end{figure}

A desirable property of the global latent space is that it is easily extensible when new models are added to the pool -- without necessarily having to retrain the adapters. In this section, we demonstrate that our framework possesses this property. We use the experiment in Section \ref{section:diff_seeds} as a scaffold and introduce a 4th model (\texttt{Seed-4}) trained with a different seed. We learn the pairs $\big(\mathbf{T}^{[\texttt{Seed-4} \rightarrow \mathbf{\Sigma}]}, \mathbf{T}^{[\mathbf{\Sigma} \rightarrow \texttt{Seed-4}]}\big)$ whilst keeping the global latent-space adapters of \texttt{Seed-\{1, 2, 3\}} fixed. 

Our results are shown in Figure \ref{fig:pool_extension}. Note that during training, we only use the paths involving \texttt{Seed-\{2, 3\}} to learn the latent-space adapters for  \texttt{Seed-4}. Though we never train on the paths $\big(\texttt{Seed-4} \rightarrow \texttt{Seed-1}\big)$ and $\big(\texttt{Seed-1} \rightarrow \texttt{Seed-4}\big)$, we are able to zero-shot translate between the two models, achieving comparable performance  relative to paths that were explicitly trained on. This demonstrates that the original latent space learned using \texttt{Seed-\{1, 2, 3\}} is indeed global -- cache blocks from different models translated into the shared latent-space are reasonably indistinguishable from each other, thus allowing arbitrary source-target translation paths once a new model learns adapters into the space. Generalization to unseen paths would not be possible if the shared latent-space was path-wise segmented or delineated by model. 

Our results in this section also demonstrate that as the size of the pool grows, we may not need all the existing models in the pool to learn the translator for incoming models. This reduces the memory and compute overhead of extending the pool to new models.

\section{A global k-v latent space enables module portability}
An implication of having a shared k-v latent-space is that learned task specific modules that manifest as k-v caches (like prefix-tuning \cite{li2021prefix}; and soft-prompts \cite{lester2021power}) can become a shared resource across models. A module trained for one model to perform a specific task can be translated into the k-v cache space of any of the other models in the pool, making those task specific skills accessible to all models without per-model training. 

Such module portability is desirable for several reasons. First, the computational cost of learning a skill is amortized across all the models in the pool. We do not have to train a module for each model since we learn the module once and then translate it into the k-v cache space of the remaining models when needed. The saved compute can be redirected to expand the pool of skills instead. Next, module portability enhances data privacy. Only one model needs to be directly exposed to the task data but the resulting module can be used by all other models. Stronger privacy guarantees can be achieved if the original module is trained via differential privacy \citep{dwork2006differential, abadi2016deep}.
\begin{figure*}[ht!]
    \centering
    \begin{subfigure}[t]{0.48\textwidth}
        \includegraphics[scale=0.28]{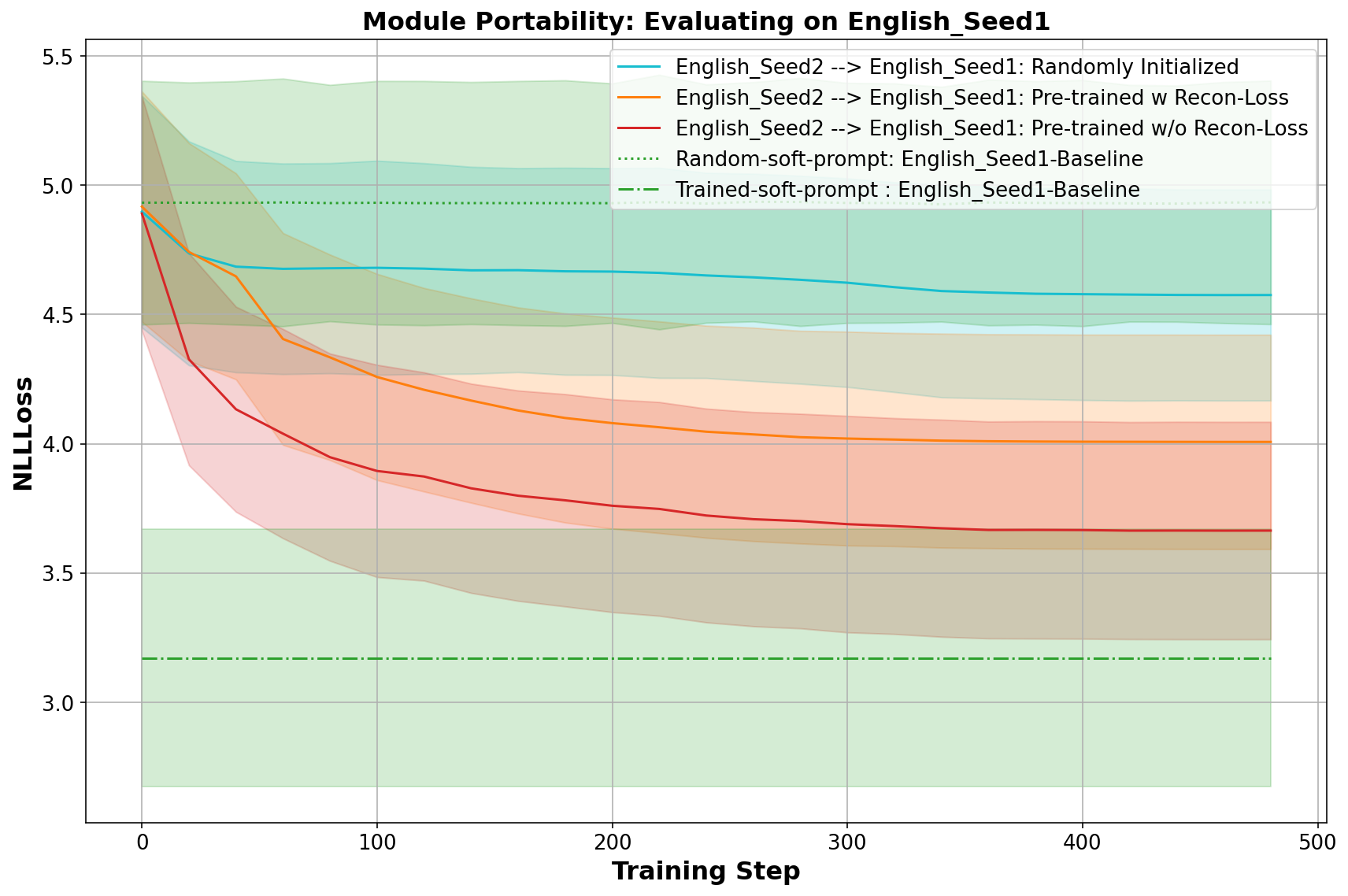}
    \end{subfigure}
    \hfill
    \begin{subfigure}[t]{0.48\textwidth}
        \includegraphics[scale=0.28]{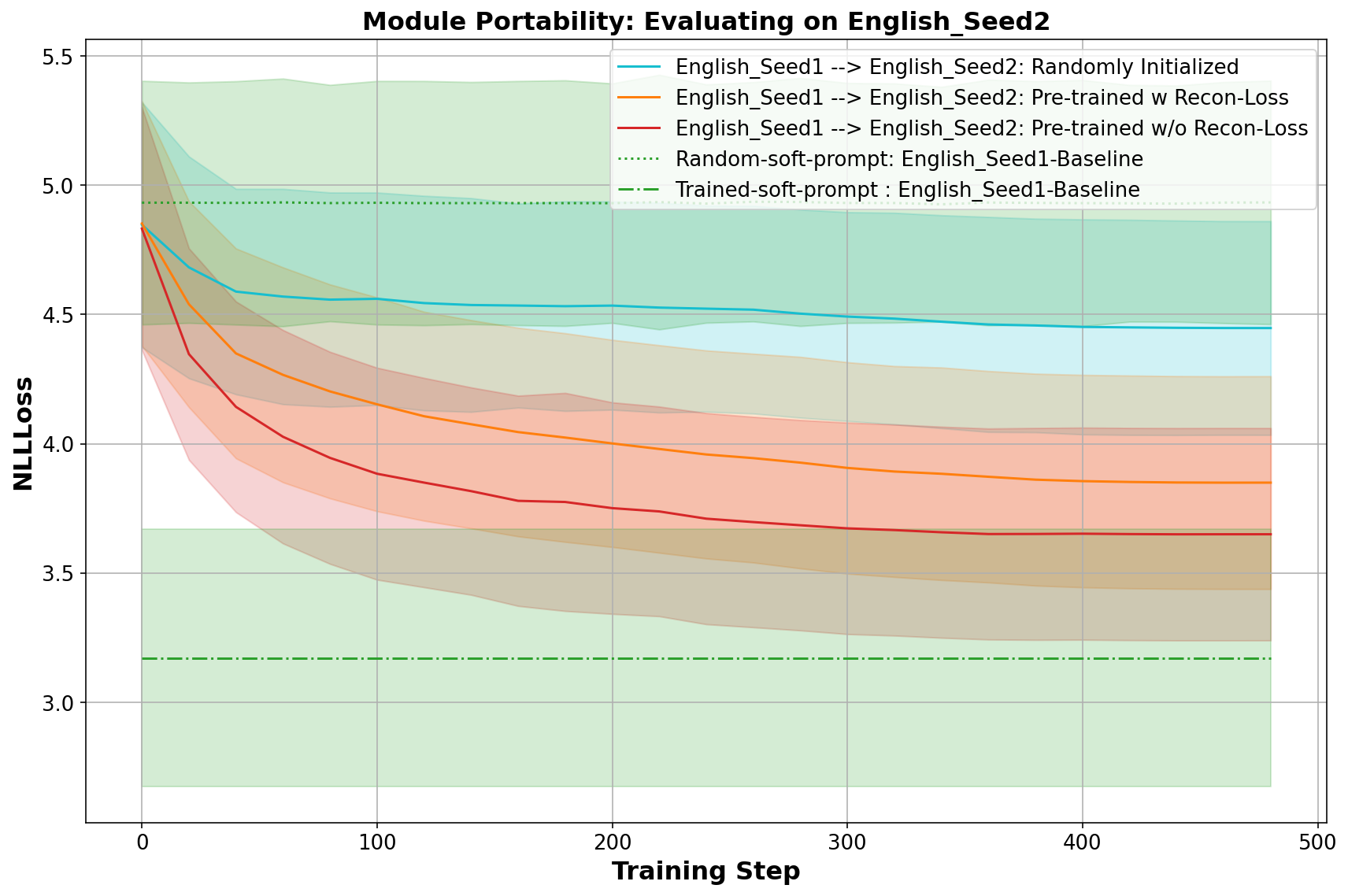}
    \end{subfigure}
    \caption{We adapt the shared global space between two models trained on different seeds to allow for zero-shot module portability. Shaded areas correspond to standard deviation over the 200 meta-eval task. We compare using pre-trained latent space adapters versus randomly initialized ones (cyan). }
    \label{fig:module_port}
\end{figure*}

\subsection{Setup}
We investigate module portability by running meta-learning experiments on the prompt recovery task. The prompt recovery task is defined as follows: given a list of k sequences of text, $\{x_1 \ldots x_k\}$, corresponding to completions from a prompted LLM: $x_i \sim m(\cdot ~|~ \mathbf{C})$, we wish to recover the context prompt, $\mathbf{C}$ by learning a soft prompt \citep{lester2021power}, $\mathbf{P} \in \mathbb{R}^{p \times D}$ such that $m(\cdot ~|~ [\mathbf{P}]) \approx m(\cdot ~|~ \mathbf{C})$. $\mathbf{P}$ should generalize to a held out set of example completions as measured by likelihood under the LLM. Since we learn $\mathbf{P}$ without seeing $\mathbf{C}$, we are learning to recover the original prompt, but as soft-tokens, instead of text. The prompt completion task is friendly to investigating module portability since we can easily create many task instances, $(\mathbf{C}_{j}, \{x_1 \ldots x_k\}_{j})$, for training the shared latent-space.

We create a seed set of 65 human written prompts \footnote{See Appendix \ref{appendix:prompt_egs} for examples} which we use to prompt Gemini-2.5 Flash to generate $\sim$ 6000 similar prompts. We specify that the generated prompts should be diverse and span multiple domains. For each of these 6k prompts, we use an instruction tuned \citep{wei2021finetuned} Gemma-2 model to generate 300-800 possible completions of length 30-50 words of which 50 completions are used for evaluations. We use 5.8K prompt completion tasks as meta-train and 200 tasks as the meta-test set. 

We use the meta-train task to train (or fine-tune) the latent-space adapters to be able to translate the k-v cache of the learned soft-prompt from the source to the target model. Given a translated cache, we evaluate its loss on the validation set on the target model (i.e zero-shot) and this loss is used as learning signal for updating the adapters into the shared global latent space.  At test time, we freeze all the translation and base-model parameters, learn a soft-prompt for each of the 200 meta-eval completion task on a source model and evaluate the performance after zero-shot translating it into the latent-space of the target model.

\subsection{Results}
\looseness=-1 Figure \ref{fig:module_port} shows that we can learn to zero-shot port modules that have been trained on one model to be used by another model. Our approach performs significantly better than the lower bound of using a random (untrained) soft-prompt per-task on the target model. With only a few thousand meta-train tasks, our approaches the upper bound performance of learning the soft-prompt directly for each target model. We posit that with more data, we can match and possibly exceed this baseline performance since our experiments featured in Figure \ref{fig:variations} show that the shared global-space can learn to generalize beyond the base model performance.

Note from Figure \ref{fig:module_port} that it is beneficial to pre-train the global latent space before fine-tuning on the meta-train set. We get the best performance by pre-training on the suffix language modelling task without the reconstruction loss.
\begin{table*}[t!]
\centering
\begin{tabular}{|l|l|l|l|l|l|l|l|l|}
\hline
& \textbf{Eval Loss} & \textbf{Identity} & \multicolumn{2}{l|}{\textbf{Linear}} &\multicolumn{4}{l|}{\textbf{Our Translator (Best HP)}} \\ \cline{4-9}
& & & \textbf{269M} & \textbf{806M} & \textbf{238M} & \textbf{323M} & \textbf{476M} & \textbf{645M} \\ \hline
\textbf{Seed 1} & 3.089 & 4.941 & 3.090 & 3.092 & 3.076 & 3.068 & 3.061 & $\boldsymbol{3.056}$ \\ \hline
\textbf{Seed 2} & 3.117 & 3.967 & 3.110 & 3.111 & 3.110 & 3.105 & 3.091 & $\boldsymbol{3.076}$ \\ \hline
\end{tabular}
\caption{We experiment with 2 200M models with different random initialization. Losses reported are language modelling loss when the target model is applied to suffix text. Translator parameter counts in are the total counts across all adapters.}
\label{tab:translator_eval}
\end{table*}

\section{Ablations}
\subsection{Performance improves with shared space adapter model size}
We ablate the use of the multi-layer cross-attention architecture introduced in Section \ref{subsec:trans_arch}. We compare to an identity and linear mapping as baselines. For the linear mapping, we have separate maps for k- and v-caches respectively which are applied to the layer-wise concatenated k-v caches. 

The results in Table \ref{tab:translator_eval} show that using an identity mapping (no learned parameters) yields poor performance even though the two models differ only in their random initialization. This mapping has the further disadvantage that it can only be used when the models' k-v cache dimensions match exactly. It is interesting to see that that we match or exceed the base model performance using a simple linear map. This is in line with recent work like \cite{moschella2022relative, pmlr-v243-lahner24a} that suggest that the geometries of latent spaces of deep neural networks trained on the same data distribution tend to be approximately equal modulo linear transformations. A caveat of the results in Table \ref{tab:translator_eval} is that that the linear maps significantly expand the size of the k-v cache dimension to create the shared latent-space (upward projections of 8$\times$ and 24$\times$ respectively for the 269M and 806M models).

Our translator outperforms the simple linear map with fewer parameters. Unlike the linear map, our cross-attention architecture exhibits desirable scaling behavior: consistent improvement in performance as we increase the size of the latent-space adapters.

\begin{figure}[h!]
\includegraphics[scale=0.43]{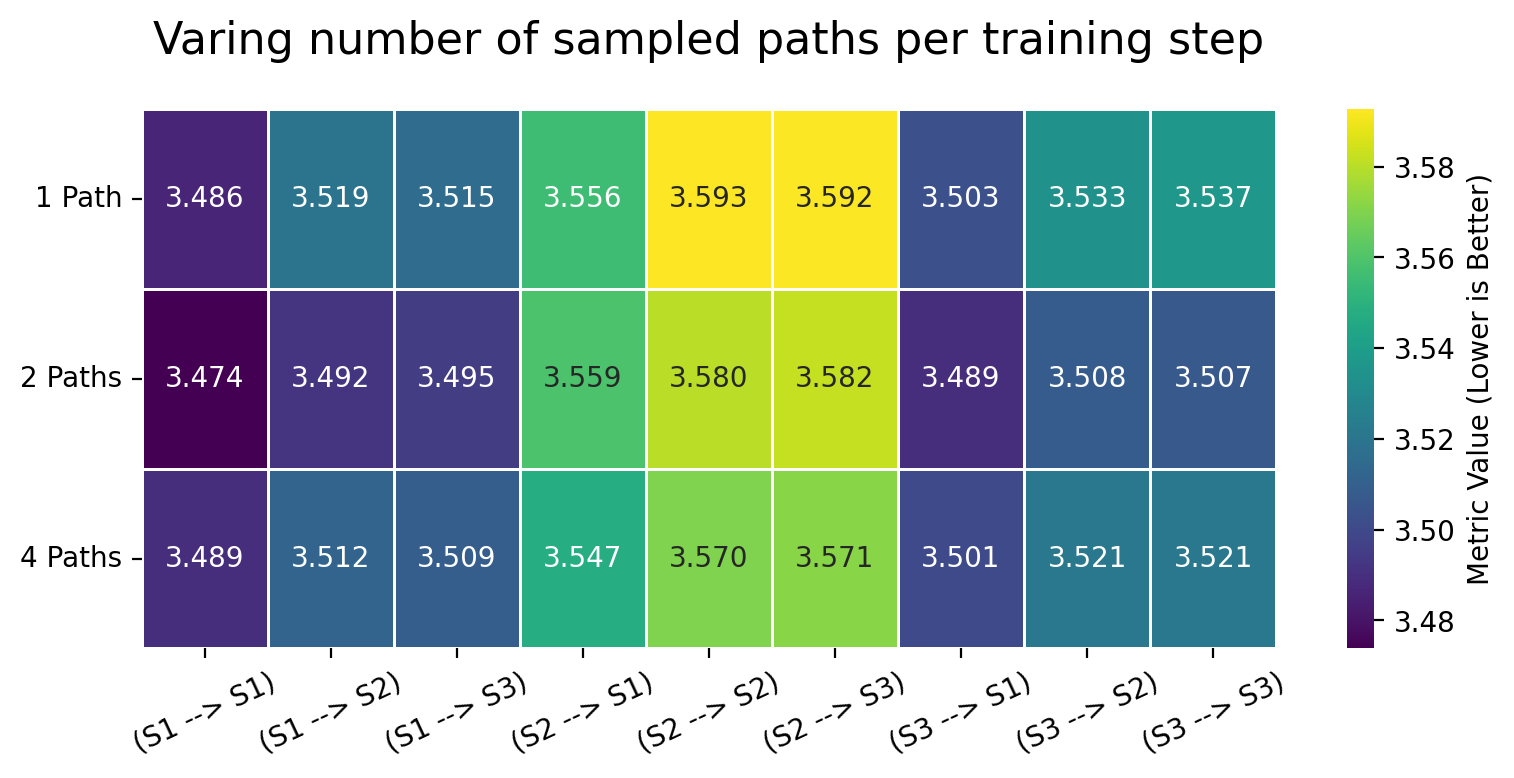}
\caption{We can learn to align latent-spaces without using all possible paths per-step during training. More paths per-step is not unilaterally better. For some translation directions, having 2 paths results in superior performance over 4.}
\label{fig:vary_num_samples}
\centering
\end{figure}
\subsection{We do not need to see all all paths in a single batch during training}
For the learning objectives proposed in Section \ref{subsection:learning_algo}, we sum over multiple paths through the latent space to compute the loss. The number of possible paths is combinatorial in the number of models in the pool. Thus, it would be computationally inadvisable, to consider all possible paths during each step when training the shared latent space. 

For all our experiments with 2 models, we use all the possible 4 paths during training. However, instantiating 9 paths for training the latent space for 3 models would be both slow and memory intensive. We use the setup from Section \ref{section:diff_seeds}  and experiment with the impact of using significantly fewer paths than possible for training. Figure \ref{fig:vary_num_samples} shows the results of our experiments -- each point is the minimum over a set of hyper-parameter sweeps. 

The average base performance of the 3 seed models is 3.460 NLL (negative log-likelihood) on the evaluation set. Even with 1 translation path per step, we can learn a map that brings us close to this baseline when we average over all 9 possible paths (3.53 NLL). When translating from $\texttt{Seed-1,3}$ to other models, we get the best performance using 2 paths per batch whilst 4 paths work best for translating from $\texttt{Seed-2}$ to other models. We posit that some amount of stochasticity is important to achieve good generalization of the learned latent space.


\subsection{Reconstruction loss is unnecessary in the presence of the suffix language modelling loss}
\begin{table}[h]
\centering
\resizebox{\columnwidth}{!}{
    \begin{tabular}{|l|l|l|}
    \hline
    & English-Seed1 $\rightarrow$ & English-Seed2 $\rightarrow$  \\
    & English-Seed2 & English-Seed1 \\ \hline 
    \textbf{Recon Weight = 0.0} & $\boldsymbol{3.406}$ &  $\boldsymbol{3.411}$ \\ \hline
    \textbf{Recon Weight = 0.5} & 3.531 & 3.501  \\ \hline
    \textbf{Recon Weight = 1.0} & 3.626 & 3.534 \\ \hline
    \end{tabular}
}
\caption{At fixed data, using the reconstruction loss did not improve learning efficiency}
\label{tab:reconstruction}
\end{table}

Many works in the space of latent space alignment use the reconstruction loss as a primary objective \citep{pmlr-v243-lahner24a, jha2025harnessing}. We however find that the presence of the target task objective (in our case, language modelling loss), obviates the need for the reconstruction loss.  Thus, for all our primary results, we excluded the reconstruction loss unless otherwise explicitly stated.
\section{Related Work}
Many areas of machine learning are concerned with solving complex problems by enabling effective communication between groups of models. Recent works in using LLMs as agents feature component models communicating with each at a high level of abstraction via natural language or structured data formats like JSON or query language \citep{wang2024openhands, wang2023interactive}. In the fields of multi-agent reinforcement learning \citep{gronauer2022multi} and emergent communication \citep{lazaridou2020emergent}, the messages passed between models usually take the form of low dimensional vectors that connect the output of one model to the input of another \citep{wu2025dense, lo2023cheap, chaabouni2022emergent, foerster2016learning}. Our framework provides higher bandwidth communication channels and facilitates much stronger model collaboration by allowing models to directly access each others' internal state in the form of the k-v cache latent spaces.

Our framework also has implications for coarse-grained mixture of expert architectures like DiPaCo \citep{douillard2024dipaco} and Small-Talk \citep{filippova2024no} where text generation proceeds by periodically switching between a set of expert models. These architectures tend to incur high inference latency due to the need to re-compute the prefix k-v caches whenever the router switches between models. By learning to translate between the k-v cache spaces of the models in the ensemble, we can significantly reduce the cost associated with naive recomputation of the prefix cache per model.

Our research builds upon the foundational work in latent space alignment, which seeks to bridge the semantic gap between disparate, pre-trained models \citep{maiorca2023latent, jha2025harnessing}. \citet{pmlr-v243-lahner24a} demonstrate that direct linear transformations can effectively map one model's latent space to another's, a finding that makes the prospect of inter-model communication computationally feasible. Similarly, work on relative representations by \citet{moschella2022relative} provides a powerful method for zero-shot composition, allowing modules from different models to be combined without fine-tuning by creating an invariant representational format. The techniques in this area have largely been applied to vision models and simpler architectures like auto-encoders with the aim of statically "stitching" together components from different models.  Our work proposes a different motivation for alignment -- enabling dynamic, stateful collaboration between models. 

\section{Conclusion}
In this paper, we presented a scalable and effective method for enabling direct latent space communication between large language models by learning a shared k-v cache representation space. This approach not only bridges the gap between incompatible models trained under various conditions but also unlocks new benefits, including improved performance and the ability to transfer learned skills between models in a zero-shot manner.

This work opens up many exciting avenues for future research. An immediate next step is to scale our framework to a larger, more heterogeneous pool of models, including those from architectural families vastly different from those explored in this work. While this work focused on language modeling as a proof of concept, future work should also explore more complex downstream tasks like agentic research \citep{lu2024ai}  and software engineering \citep{wang2024openhands}. Ultimately, this line of inquiry moves us closer to a paradigm of seamless inter-model collaboration, where the collective capabilities of diverse agents can be harnessed to solve problems of unprecedented complexity.

\bibliography{main}
\newpage
\onecolumn

\appendix
\section{Model Details}

\begin{table}[ht]
\centering
\caption{Overview of model parameters and design choices.}
\label{tab:model_params}
\begin{tabular}{@{}lccc@{}}
\toprule
\textbf{Parameters}   & \textbf{100M}    & \textbf{200M}    & \textbf{400M}   \\ \midrule
$d_{\text{model}}$    & 960           & 960           & 960           \\
Layers                & 4             & 8             & 16             \\
Pre-norm              & yes            & yes            & yes            \\
Post-norm             & yes            & yes            & yes            \\ \midrule
Non-linearity         & GeGLU          & GeGLU          & GeGLU          \\
Feedforward dim       & 15360          & 15360          & 15360          \\ \midrule
Head type             & GQA            & GQA            & GQA            \\
Num heads             & 4              & 4             & 4            \\
Num KV heads          & 1              & 1              & 1             \\
Tied embedding        & yes            & yes            & yes            \\ \bottomrule
\end{tabular}
\end{table}

\section{Prompt Examples}
\label{appendix:prompt_egs}
Here are some example prompts we used for the prompt completion task.
\begin{itemize}
    \item \texttt{tell me a story about how the duck and the rabbit got angry with each other, but then later realized it was all a misunderstanding.}
    \item \texttt{can you teach me how to solve systems of linear equations? give me lots of examples.}
    \item \texttt{describe for me the history of the three kingdoms period in Chinese history.}
    \item \texttt{describe an excercise routine for dogs that could be used as a soft-drink ad.}
    \item \texttt{write a description of a disaster movie in the form of a cake recipe.}
    \item \texttt{tell me knock knock jokes, but make sure they are not funny.}
    \item \texttt{start by telling a childrens story about a dog and a cat but it seemlessly becomes a spy thriller.}
    \item \texttt{list all the prime ministers of the UK but encode them with a ceaser cipher.}
    \item \texttt{give me a treatment for an Eat Pray Love sequal Eat2Pray2Love2.}
    \item \texttt{write the code for red black trees first in python, then c\# then in a very obscure language.}
    \item \texttt{write a musical about Demis Hassabis.}
    \item \texttt{pick one of the three stooges from the three stooges and write a poem about them.}
    \item \texttt{roll a 20 sided die and tell me the result.}
    \item \texttt{generate me a dungeons and dragons character for me, but cheat and make the stats slightly too good.}
    \item \texttt{write an essay about Robert Moses but make it kind of mediocre and have at least 2 mistakes in it.}
    \item \texttt{assume 1+1=3 and write a program to solve the quadratic equation x\^{}2-3x+2=0.}
    \item \texttt{you are a pteridactl who is trying to convince a friend to buy a car.}
\end{itemize}

\section{Overflow Experimental Results}

\begin{table}[h]
\centering
\begin{tabular}{|l|l|l|}
\hline
& English-Seed1 $\rightarrow$ & English-Seed2 $\rightarrow$  \\
& English-Seed2 & English-Seed1 \\ \hline 
Random Soft-Prompt & $4.932_{0.471}$ & $4.932_{0.471}$ \\ \hline
Randomly Initialized  & $4.847_{0.474}$ & $4.897_{0.448}$ \\ \hline
Pre-trained w Recon-Loss & $4.852_{0.473}$ & $4.917_{0.445}$ \\ \hline
Pre-trained w/o Recon-Loss & $4.832_{0.469}$ & $4.890_{0.453}$ \\ \hline
\end{tabular}
\caption{Zero-shot performance on module portability task \textbf{before meta-training}. Meta-training is key to adapting our framework for module portability.}
\label{tab:before_meta-training}
\end{table}
\end{document}